\documentclass[fleqn,10pt]{wlscirep}
\usepackage[utf8]{inputenc}
\usepackage[T1]{fontenc}
\usepackage{textcomp}
\usepackage[para,online,flushleft]{threeparttable}

\title{Crop Yield Prediction Integrating Genotype and Weather Variables Using Deep Learning}

\author[a,1]{Johnathon Shook}
\author[b,1]{Tryambak Gangopadhyay}
\author[b]{Linjiang Wu}
\author[b]{Baskar Ganapathysubramanian}
\author[b, 2]{Soumik Sarkar}
\author[a, 2]{Asheesh K. Singh}

\affil[a]{Department of Agronomy, Iowa State University, Ames, IA, 50011}
\affil[b]{Department of Mechanical Engineering, Iowa State University, Ames, IA, 50011}

\affil[1]{J.S. and T.G. contributed equally to this work.}

\affil[2]{To whom correspondence may be addressed. Email: singhak@iastate.edu, soumiks@iastate.edu.}

\begin{abstract}
Accurate prediction of crop yield supported by scientific and domain-relevant insights, can help improve agricultural breeding, provide monitoring across diverse climatic conditions and thereby protect against climatic challenges to crop production including erratic rainfall and temperature variations. 
We used historical performance records from Uniform Soybean Tests (UST) in North America spanning 13 years of data to build a Long Short Term Memory - Recurrent Neural Network based model to dissect and predict genotype response in multiple-environments by leveraging pedigree relatedness measures along with weekly weather parameters.  
Additionally, for providing explainability of the important time-windows in the growing season, we developed a model based on temporal attention mechanism.
The combination of these two models outperformed random forest (RF), LASSO regression and the data-driven USDA model for yield prediction. 
We deployed this deep learning framework as a 'hypotheses generation tool' to unravel GxExM relationships. 
Attention-based time series models provide a significant advancement in interpretability of yield prediction models. 
The insights provided by explainable models are applicable in understanding how plant breeding programs can adapt their approaches for global climate change, for example identification of superior varieties for commercial release, intelligent sampling of testing environments in variety development, and integrating weather parameters for a targeted breeding approach. Using DL models as hypothesis generation tools will enable development of varieties with plasticity response in variable climatic conditions.  We envision broad applicability of this approach (via conducting sensitivity analysis and "what-if" scenarios) for soybean and other crop species under different climatic conditions. 

\end{abstract}
\begin{document}

\flushbottom
\maketitle
%
%
\thispagestyle{empty}

\vspace{-0.5 cm}


\section*{Introduction}
One of the key challenges in plant breeding and crop production is to predict performance (seed yield) in unseen and new environments.
This active research area is complicated by the time and expense of generating an extensive dataset to represent a wide range of genotypes and environments. 
Among different crops, soybean has a long history of cultivation in North America, with the first reported production in Georgia in 1766 \cite{hymowitz1983introduction}. 
Over the years, production in the US and Canada has expanded longitudinally as far west as Kansas-Colorado border and latitudinally from southern Texas to Canada \cite{websitereference_1,websitereference_2}.
North American annual soybean yield trials (known as Uniform Soybean Tests (UST)) have been coordinated in the United States and Canada through the United States Department of Agriculture (USDA) between public breeders in university and government settings since 1941 \cite{websitereference_4, websitereference_5}. These trials are used to evaluate current and experimental varieties in multiple environments within their range of adaptation. Therefore, these trials are valuable sources of historical and current data to improve prediction performance with the assimilation of genetic and environmental variables. Management and permanent environmental effects have been examined primarily at small scales due to the labor required for managing large numbers of plots \cite{zhang2016warming, puteh2013soybean}.
With the addition of each layer of added characterization of the environment, less of the differences need be ascribed to a generic "environmental" component, and can instead be examined individually in combination with plant genetics. The nexus of genetic and non-genetic variables form the cornerstone of plant breeding strategies, irrespective of crop species, for meeting crop production challenges in the future\cite{lenaerts2019improving,Wulff2019breed}. 

Climatic resiliency in cultivars is an important objective for plant breeders and farmers to get a high seed yield in a myriad of environments\cite{ICARDA2018resil}. The climatic variability can be associated with changes in temperature and rainfall events (including patterns and magnitude) and other weather variables. 
In addition to spatial variability, temporal variability of weather variables \cite{websitereference_3} is equally important but generally less understood or not included in yield prediction studies.
It is important to understand how agricultural production is affected by the variability of weather parameters in presence of global climate change, especially with higher occurrence of extreme weather events. Therefore, prediction of the effects of changing environments on performance can help in making informed plant breeding decisions, marketing decisions, optimizing production and comparing results over multiple years \cite{jagtap2002adaptation}.

Traditionally, crop growth models have been proposed to simulate and predict crop production in different scenarios including climate, genotype, soil properties, and management factors~\cite{blanc2017statistical}. 
These provide a reasonable explanation on biophysical mechanisms and responses but have deficiencies related to input parameter estimation and prediction in complex and unforeseen circumstances~\cite{roberts2017comparing}. 
Previous attempts at yield prediction across environments have relied on crop models generated by quantifying response in a limited number of lines while altering a single environmental variable, limiting the inference scope ~\cite{bishop2014seasonal}. 
To bypass the limitations of crop growth models, linear models have also been used to predict yield with some success ~\cite{jewison2013USDA}. However, these low-capacity models typically rely on a rather small subset of factors, therefore failing to capture the complexity of biological interactions and more site-specific weather variable complexities. 
Traditional linear methods such as Autoregressive Integrated Moving Average (ARIMA) have been used for time series forecasting problems \cite{petricua2016limitation}, but these methods are effective in predicting future steps in the same time-series. For time series prediction tasks, deep neural networks show robustness to noisy inputs and also have the capability to approximate arbitrary non-linear functions~\cite{dorffner1996neural}. Deep learning models can provide solutions in the presence of such complex data comprising of different weather variables, maturity groups and zones, and genotype information. 

Long Short Term Memory (LSTM) networks are very useful for time series modeling as they can capture the long-term temporal dependencies in complex multivariate sequences \cite{malhotra2015long}. 
LSTMs have shown state-of-the-art results in various applications including off-line handwriting recognition \cite{doetsch2014fast}, natural language processing \cite{sutskever2014sequence} and engineering systems \cite{gangopadhyay2020deep}. 
LSTMs have also been used effectively for multivariate time series prediction tasks \cite{jiang2018predicting,gangopadhyay2018temporal, shook2018integrating}.
Considering the importance of climate extremes for agricultural predictions, random forest has been utilized to predict grid-cell anomalies-deviations of yields \cite{vogel2019effects}.
Previous work \cite{you2017deep} using deep learning for yield prediction has utilized multi-spectral images to predict yield (instead of leveraging only multivariate time series as input) without considering model interpretability.
Khaki et al. \cite{khaki2019crop} applied deep neural networks for yield prediction of maize hybrids using environmental data, but their model is not capable of explicitly capturing the temporal correlations and also
lacks explainability.
LSTM based model has been used for corn yield estimation \cite{jiang2019deep}, but these models lack interpretability. This study is based on geospatial data without field-scale farming management data and lacks temporal resolution in the absence of daily weather data. 
Attention based LSTM has been used along with multi-task learning (MTL) output layers \cite{lin2020deepcropnet} for county level corn yield anomaly prediction only based on meteorological data (maximum daily temperature, minimum daily temperature) without field-scale farming data. 
Other approaches to predict yield rely on the use of sensors to identify the most informative set of variables to predict yield\cite{parmley2019tpp,parmley2019scirep}, which is very useful in multiple applications; however, there is still a need to integrate weather parameters and in a time series approach involving multiple genotypes.
Using these motivations, we developed a model that can capture the temporal variability of different weather variables across the growing season in an explainable manner to predict soybean yield from the UST dataset of field trials spanning 13 years across 28 states and provinces.

\begin{figure*}[tbhp]
\begin{center}
\setlength{\unitlength}{0.012500in}%
\includegraphics[width=12cm, height=8 cm]{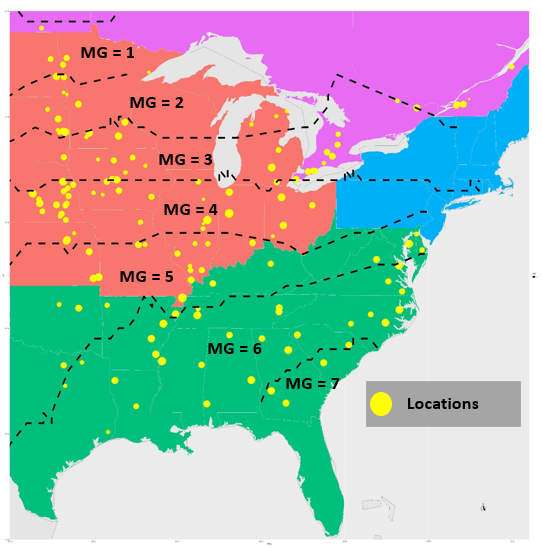}
\end{center}
\caption{Map showing different locations in the USA and Canada included in our dataset. The dataset comprises of different maturity groups (MGs), some of which are labeled in the figure. The relative size of a yellow dot (representing location) indicates the size of the dataset for that particular location. Dataset included observations from the National Uniform Soybean Tests for years 2003-2015 and is split into North (MG 0 to 4) and South (MG 4 to 8) regions\cite{UST2018S, UST2018N}, consisting of 103,365 performance records over 13 years and 150 locations. These records are matched to weekly weather data for each location throughout the growing season (30 weeks). This generated a dataset with 35,000 plots having phenotype data for all agronomic traits.}
\label{map_details} 
\end{figure*}




We propose a framework based on LSTM and temporal attention to predict crop yield with 30 weeks of weather data per year (over 13 years) provided as input, along with a reduced representation of the pedigree to capture differences in the response of varieties to the environment.
We vary the number of input time-steps and compare the performance of our proposed Temporal Attention model with the Stacked LSTM model for two variations of each model.
We also compared against the results of random forest (RF), LASSO regression and the data-driven state-of-the-art USDA model.
The temporal attention mechanism highlights the significant time periods during the growing season leading to high or low yield prediction, concurred with domain knowledge.
In this paper, we report improved fidelity interpretation of the prediction outcomes without sacrificing the accuracy for multivariate time-series prediction.
Our proposed framework can have widespread applications in plant breeding, crop science research, and agricultural production.


\section*{Methods}

\subsection*{Preparation of Performance Records}
Files from 2003-2015 USTs were downloaded as PDFs \cite{websitereference_4, websitereference_5}. 
Using on-line utility Zamzar (zamzar.com), all 26 PDFs from this period were converted to .xlsx files, with each tab corresponding to a single page in the file. 
In this way, the vast majority of tables were recovered with no errors or need for human translation. However, random checking for error was manually performed to ensure verity. 
These tables were manually curated to align all performance records for a given genotype/location combination into a single row. 
Records that did not have yield data (due to a variety not being planted in a specific location or dying prior to production of seed), were removed from the file.

Following data cleaning, the final dataset comprised of 103,365 performance records over 13 years representing 5839 unique genotypes, along with all available management information. After compilation, we imported performance records in Python for further data analysis. 

\subsection*{Acquisition and Sub-Sampling of Weather Records}
Daily weather records for all location/year combinations were compiled based on the nearest available weather station (25km grid) on Weather.com. We downsampled the dataset to include maximum, minimum, and average conditions on different time frames throughout the growing season (defined April 1 through October 31) and this information was appended to performance records. 


\subsection*{Genotype Clustering}

We included genotype-specific criteria to apply the model for specific genotypes and mean location yield across genotypes. 
Due to the nature of the UST program, most of the genotypes tested in this period do not have molecular marker data available, preventing the use of a G matrix. To circumvent these restrictions, we developed a completely connected pedigree for all lines with available parentage information, resulting in the formation of a 5839 x 5839 correlation matrix. To improve the model performance, genotypes were clustered based on the organization which developed them, providing additional control over relatedness. 

We clustered genotypes in 5 clusters using the K-means Clustering technique based on the correlation matrix to extract information about relatedness. With a specified number of clusters ($n$), the K-means algorithm finds $n$ groups of equal variance by choosing centroids of the clusters to minimize a criterion known as $inertia$ (also called, within-cluster sum-of-squares). This algorithm is effective for a large number of samples and finds application across different domains. With this hard clustering technique, each genotype belongs to one of the 5 clusters. The clustering is used to represent each line as a function of membership in 5 groups, which is fed into the model to allow differentiation of lines.




\subsection*{Model Development}
To leverage the temporal sequence of variables, a modeling approach based on recurrent neural network (RNN) was developed to capture correlation across time.
Gradient descent of an error criterion may be inadequate to train RNNs especially for tasks involving long-term dependencies \cite{bengio1994learning}.
To overcome these challenges, long short-term memory (LSTM) was used, which is an RNN architecture designed to overcome the error backflow problems \cite{hochreiter1997long}. 
By learning long-range correlations in a sequence, LSTM can accurately model complex multivariate sequences~\cite{malhotra2015long}.



\begin{figure*}
\begin{center}
\setlength{\unitlength}{0.012500in}%
\includegraphics[width=16cm, keepaspectratio]{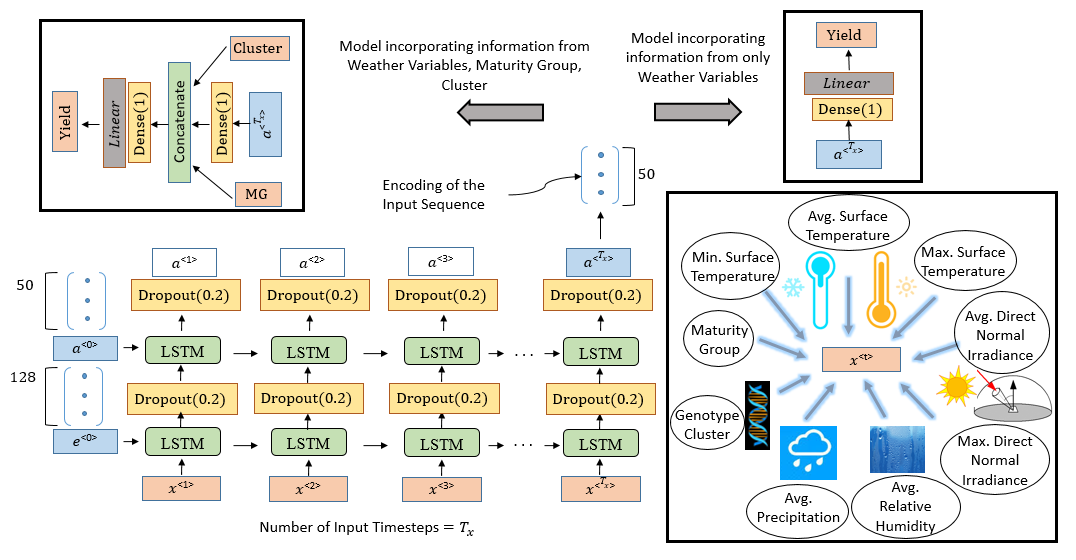}
\end{center}
\caption{The figure showing the Stacked LSTM Model. The input feature vector is $x^{<t>}$ at time-step 't'. Depending on whether the maturity group and genotype cluster information are incorporated in the model or not, the vector $x^{<t>}$ can be 9-dimensional or 7-dimensional. We included 7 weather variables in our study. The embedding vector $a^{<T_x>}$ encodes the entire input sequence and summarizes the sequential dependencies from the time-step 0 to the time-step $T_x$. We designed two variants of our proposed model based on input information with the time series encoding part remaining the same for both variants. This model (when including MG, cluster) had 106,511 learnable parameters and the training time/epoch was 60 secs.}
\label{stacked_lstms_model} 
\end{figure*}

\begin{figure*}
\begin{center}
\setlength{\unitlength}{0.012500in}%
\includegraphics[width=13cm, keepaspectratio]{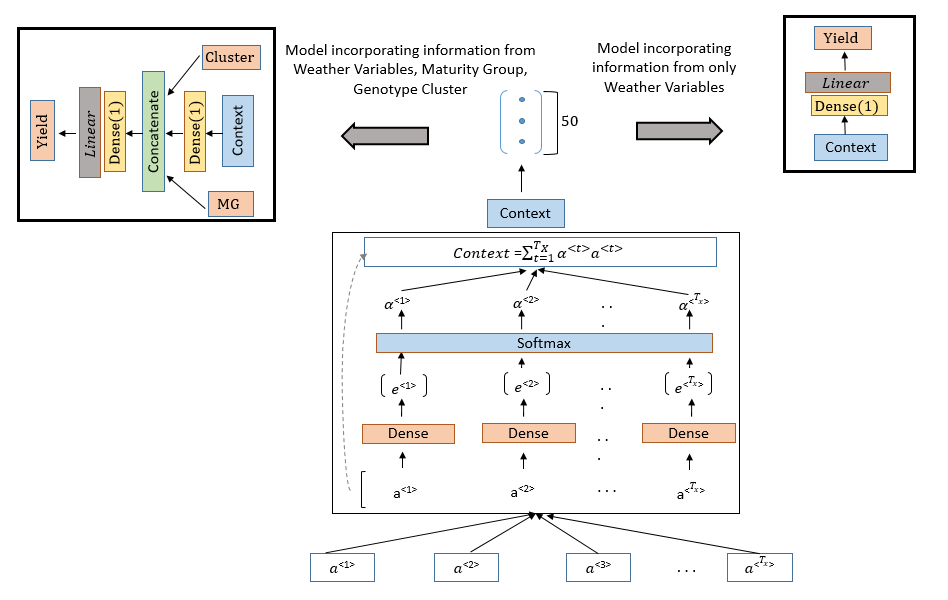}
\end{center}
\caption{The figure showing the Temporal Attention Model. The LSTM encoding part is the same as that of the Stacked LSTM Model where we get the annotations $a^{<t>}$ for each timestep. Instead of only using $a^{<T_x>}$, this model utilizes all annotations which act as inputs for the temporal attention mechanism. Based on the computed context vector, the two variants of this model are designed depending on the input information. This model (when including MG, cluster) had 106,562 learnable parameters and the training time/epoch was 60 secs.}
\label{temporal_attention_model} 
\end{figure*}

We developed two models, based on LSTM: (a) Stacked LSTM Model (without using any attention) (Fig.~\ref{stacked_lstms_model}), and (b) Temporal Attention Model (using a temporal attention mechanism) (Fig.~\ref{temporal_attention_model}). The output of both the models is yearly seed yield as this is a many-to-one prediction problem. For each model, we formulated the model variants depending on whether the performance records comprise data of maturity group and genotype cluster. The same modeling approach was used to compute the time-step wise encoding for both models. 
Two stacked LSTM layers were used to encode the $T_x$ time-steps of the input sequence as shown in Fig.~\ref{stacked_lstms_model}. Depending on the variant, for both models, we concatenated MG and genotype cluster values with the compressed time-series information. 


In the Stacked LSTM Model, the last hidden state of the encoding part is assumed to be the compressed representation from the entire input sequence. This fixed-dimensional representation was used for predicting the output value of seed yield (Fig.~\ref{stacked_lstms_model}). For the Temporal Attention Model, the compressed information (context) is computed after aggregating the information from the sequence of hidden states using the attention mechanism. 
The concept of soft temporal attention \cite{bahdanau2014neural} was first proposed in the context of neural machine translation to overcome the bottleneck of the encoder-decoder model\cite{cho2014learning, sutskever2014sequence} for long sequences. Compressing all information from the input time-steps into a fixed-length single vector was the major bottleneck for the encoder-decoder model. Temporal attention can be applied for many-to-many time series prediction \cite{gangopadhyay2018temporal} and many-to-one-prediction \cite{gangopadhyayexplainable,gangopadhyay2019deep}. The proposed approach (Fig.~\ref{temporal_attention_model}) does not incorporate a decoder LSTM as we are performing a many-to-one prediction problem. Taking in the annotations of all time-steps as input, the attention block aggregates the information and computes the context vector.   
A greedy search method was utilized to empirically determine the most influential weather variable on seed yield prediction considering data of both the northern and southern U.S. regions. In the first step of the greedy search, the Stacked LSTM model was trained for each of the 7 variables and choose the variable that had the least RMSE. With this variable added, in the second step, the model was trained for each of the other 6 variables. In this way, variables were added. More information is provided in the supplementary materials (Supplementary Tables 5, 6 and 7).  


All input features were scaled in the range (-1, 1) with the scaler fitted on the training set. We compute the Root Mean Square Error (RMSE) after inverting the applied scaling to have forecasts and the actual values in the original scale. Data were randomly split into training (80\%), validation (10\%) and test (10\%) sets. 
Models were evaluated by computing RMSE for the test set.
Both models were trained for 200 epochs to get the optimal RMSE scores. For training, Adam optimizer was used \cite{kingma2014adam} (learning rate of 0.001) and the mean squared error loss function was computed.  
Models were developed using Keras \cite{chollet2015keras} with the TensorFlow backend \cite{abadi2016tensorflow} and the models were trained using NVIDIA GPUs.


\section*{Results}
To select hyper-parameters (determination of appropriate temporal sampling of weather information to predict yield using our proposed frameworks), the test set RMSE was used to determine optimal (lowest RMSE) number of time points to predict seed yield. Using a step-wise approach building from monthly, bi-weekly, weekly and finally daily data, similar performance was observed in each scenario (approximate test RMSE = 7.206) except for daily data. The intermediate scenario of weekly data was picked for all subsequent analyses, to facilitate faster training of LSTMs and also not to downsample to a higher extent in capturing the long-range temporal dependencies. 


\begin{figure*}[tbhp]
\begin{center}
\setlength{\unitlength}{0.012500in}%
\includegraphics[width=17cm, keepaspectratio]{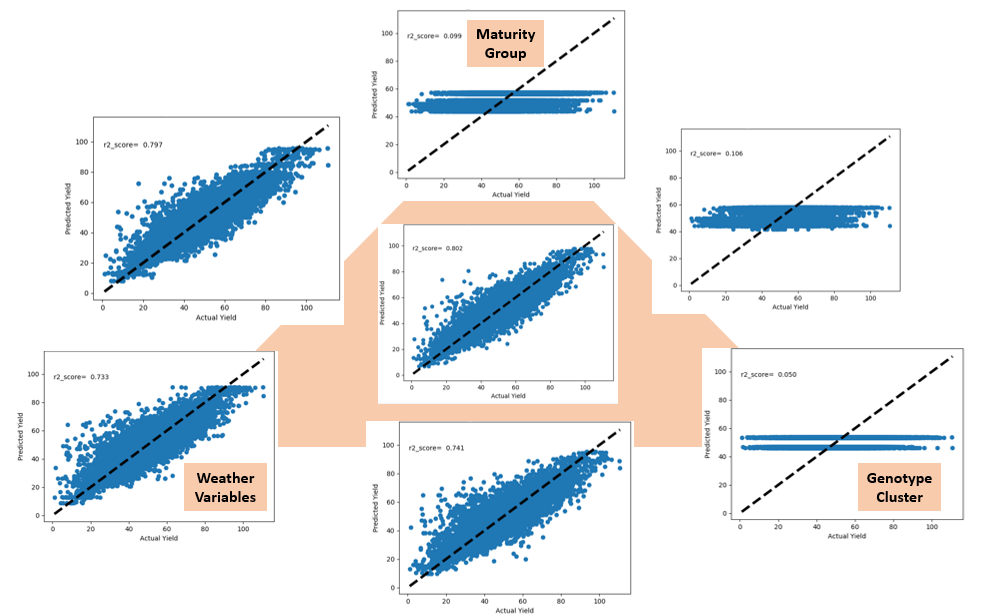}
\end{center}
\caption{Results for different inputs to the Stacked LSTM model. The vertices of the triangle demonstrate results including only the maturity group, only genotype cluster and only weather variables in the input. The edges show the results with a combination of inputs from the respective vertices. The results showed improvement when the genotype cluster was included with weather variables. The coefficient of determination increased further when the maturity group was included with weather variables. The best result were noticed when information from all sources was incorporated (shown at the center of the triangle). The best performance (RMSE = 7.130) is about 14\% of the average seed yield for the test set (50.745) and 44.5\% of the standard deviation (16.019).}
\label{prediction_results} 
\end{figure*}

Using weekly weather aggregate data in our model, the prediction models were built starting with a heuristic variable importance given to each variable.   
For example, precipitation was deemed to be most important followed by average surface temperature and so on. 
However, the largest drop in test RMSE was observed for the maturity group when it was used as a predictor factor in the model and adding the MG classification after the 2nd LSTM as well caused a further improvement in model performance. 
No perceptible change in performance was observed with variation of the number of clusters (5, 10, 15, 20, 25) using the hard clustering technique (K-means clustering). 
Therefore, subsequent analyses are done using 5 clusters in the proposed models for prediction and variable search. 
Adding the genotype cluster information at every time-step and also after the 2nd LSTM, showed better results. 


From our greedy search, we observed average relative humidity had the lowest test RMSE. 
With the inclusion of average relative humidity in the prediction model, average direct normal irradiance was the next most important variable.
Sequentially, the remaining weather variables were: maximum direct normal irradiance, maximum surface temperature, minimum surface temperature, average surface temperature, and average precipitation. 
A second greedy search initiated with the inclusion of maturity group and pedigree-based clustering revealed minimum surface temperature as the most important weather variable (lowest RMSE). 
The greedy search results revealed the following sequence of weather variable importance obtained from a forward selection approach: average direct normal irradiance, average surface temperature, maximum direct normal irradiance, average precipitation, average relative humidity, and maximum surface temperature. 
Noticeably, the ranking of the variables was different but the absolute change in RMSE scores was minimal.  

Overall, a correlation of 0.894 between predicted and observed yields in the testing and validation sets was attained; largely capturing the differences in performance between environments and years. 
However, the model remains somewhat limited in its ability to generate genotype-specific yield predictions due to the limited complexity of relationships which can be modeled using LSTM, and a lack of genomic information on each genotype. Since, a lack of molecular marker data for each line precludes us to leverage genomic prediction and its integration with the LSTM model, it is the next step of the approach presented in our paper.
As currently implemented, the model's average absolute error is 5.4 bu/acre, which is reasonable given the levels of variability within a given environment/year combination. For example, in Ames, IA, during 2003, yields ranged from 33.3-55.3 bu/acre. 
In spite of this large range of difference, an average error of only 4.5 bu/ac was observed for this environment. 
No perceptible trends are observed when we looked at state wide results combined over years. 
We also looked at originating breeding state as well as private company entries, and no geographical trends were noticeable. 

Both proposed models (Stacked LSTM, Temporal Attention) showed similar performance, and results improved when more information were included (Fig.~\ref{prediction_results}). The coefficient of determination was highest (0.802) when information from all the sources (maturity group, genotpye cluster, weather variables) were incorporated. The best model performance (test RMSE = 7.130) was \~14\% of the average yield for the test set (50.745) and 44.5\% of the standard deviation (16.019)(Fig.~\ref{prediction_results}).
Comparatively, test RMSE of 12.779 was obtained from least absolute shrinkage and selection operator (LASSO) regression, while Random Forest test RMSE was 9.889 with same input features. Therefore, both Stacked LSTM and Temporal Attention models outperform LASSO and RF models.

In comparison with the data-driven state-of-the-art USDA model\cite{jewison2013USDA}, our deep learning approach performs significantly better demonstrating much lower absolute errors. 
The USDA approach uses a linear regression approach with coefficients based on historical statewide yields and weather averages. 
However, the USDA model does not predict performance for individual locations.
Due to this limitation, we compare results of our model with the USDA model using year wise average across states for the test set.
In comparison with the USDA model, the absolute errors of our model are lower for all 12 years (except in 2011). For 2014 and 2015, the absolute errors of deep learning models were 0.03 and 0.35 (compared to 1.32 and 1.70 for the USDA model), respectively. Detailed comparison results are provided in the supplementary material (Supplementary Table 10).  

\begin{figure*}[tbhp]
\begin{center}
\setlength{\unitlength}{0.012500in}%
\includegraphics[width=17cm, keepaspectratio]{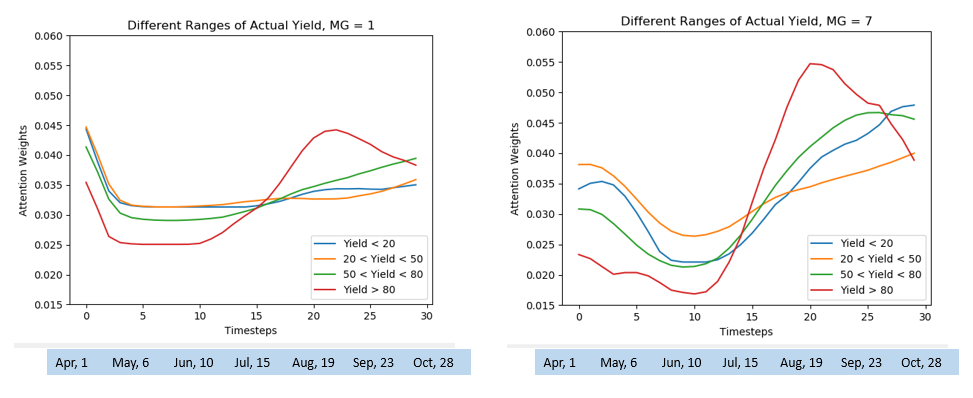}
\end{center}
\caption{Results showing the distribution of attention weights for the entire input sequence (spanning the growing season). Considering different ranges of actual yield, the results are demonstrated for two different maturity groups (MG = 1, MG = 7) providing stark geo-climatic regions (Fig.~\ref{map_details}). Early season variables were observed to be comparatively less important for prediction of the highest yielding genotypes.}
\label{attention_results} 
\end{figure*}

In addition to accurate yield prediction, the Temporal Attention Model provided insights (Fig.~\ref{attention_results}) about how early-season variables were less important for yield prediction in the highest yielding genotypes for two geographically distinct maturity groups: MG1 (Northern US adaptation) and MG7 (Southern US adaptation). We observed mild sigmoid curves for the highest yielding group in the case of both MG1 and MG7. However, we note that while MG1 had a significantly large number of plots ($\approx 550$) for the highest yielding group, MG7 had only about $30$ such plots. It points to the increasing importance of features in the August – September time phases for both North and South US regions. These time phases coincide with crop reproductive phases, emphasizing their importance in the final yield, and need functional validation which is outside of the scope of our research. However, this is an example of hypotheses generation advantage of these models motivating future research.  



\section*{Discussion}
We establish the potential for use of a long short-term memory-based method for yield prediction to allow models to account for temporal differences in the occurrence of weather events. 
Predictions using this system can be made reasonably accurate due to a large amount of training data made available through the mining of historical records. 
Our approach (using LSTM and attention) is an efficient modeling scheme to analyze soybean crop growth interaction with the weather, and to identify hypothesis for plasticity, as well as to identify key physio-environmental features that are important to include into any predictive model. 
For example, differences in the timing of extreme heat events, as well as drought periods, would affect soybean plants in various ways depending on the stage of plant development. 
For example, heat stress during flowering is particularly damaging while heat in vegetative stages of development may not produce significant reduction to harvested yield ~\cite{westgate1993flower}. 
With a larger encompassing dataset, breeders and researchers can be empowered to parse out most informative time periods, weather variables and crop responses. This information sets up the framework for breeding strategies to develop climate resilient and responsive varieties.  

Our results -- via our hypothesis generation approach -- show a potential mismatch in the heuristic/empirical results for the importance of weather variables. The finding of minimum surface temperature as the most significant weather variable suggests that nighttime temperatures play a larger role in yield prediction than previously suggested~\cite{gibson1996influence}. Our study is a retrospective design, and cannot conclude definitively that this is the case; however, these findings necessitate further empirical investigations and can be used to formulate the next set of hypotheses. Our findings are significant, as minimum temperatures have been reported to be increasing at a faster rate than maximum temperatures~\cite{karl1993asymmetric}. 
More studies are needed to ascertain the relative importance of these variables and can motivate morpho-physiological attentive breeding approaches to assemble sturdier varieties for future scenarios. 

A large capacity machine learning approach, such as the one presented in this paper using LSTM-RNN will be robust to incorporate weather changes and adjust performance predictions accordingly. 
Additional information that may improve the results of this approach is the inclusion of any supplemental irrigation provided, soil fertility levels, disease pressure and resistance levels, and direct genetic markers for the tested varieties, all of which would further strengthen predictive ability.
Therefore, future implementations may be expanded to include genomic data, additional factors such as preceding crop, row spacing, planting date, soil texture, or additional temporal data in the forms of soil sensor measurements and remote sensing data for morphological and physiological traits. The approach presented in this work will further enhance phenomic assisted breeding that collects in-season data using different sensors and payloads~\cite{parmley2019scirep,parmley2019tpp,gao2018novel} using machine and deep learning approaches suitable in plant sciences applications~\cite{singh2016machine,singh2018deep,ghosal2018explainable}.

Our work shows a unique strategy to assimilate and utilize complex data for seed yield prediction. For comparative purposes, we compared our models with the RF, LASSO and the data-driven USDA model. 
The USDA model has a limitation on the type of data it can utilize and is limited in its application. For example, as the USDA model computes predictions at the state level, the finer resolution available with our model may help in making regional marketing decisions, as well as in creating yield predictions which can capture intra-state variation due to factors such as differences in rainfall in different areas of the state. Since our results are built on more than a decade of data, it also reflects that early season weather variables are less useful in seed yield prediction and needs empirical evidence to confirm the genetic variability in plasticity of soybean genotypes in earlier stages of growth and development. Importantly, we emphasize that the utilization of the attention module within a LSTM framework allows us to tease out potentially important features for further testing. This alleviates the disadvantage of DL models -- which serve as purely blackbox predictive models -- by allowing  for hypothesis generation that will allow scientific insight via targeted follow up analysis and experiments. 


The advantages of LSTM based models have been recently established for maize yield prediction at a county level \cite{jiang2019deep}, but the model lacked interpretability. Attention based LSTM along with multi-task learning (MTL) output layers has also been used for maize yield prediction using county level data based on meteorological data (maximum daily temperature, minimum daily temperature, and daily precipitation) \cite{lin2020deepcropnet}. These studies are important for solving the yield prediction challenge; however, models are based on geospatial data without field-scale farming management data and variety information is indiscernible, and based on limited weather variables. In our soybean study, we included seven weather variables and detailed field-scale farming data with multiple maturity groups spanning continental U.S. and full variety representation.

We have shown that an LSTM-based approach can improve seed yield prediction accuracy due to the ability to identify both temporal effects of weather events and the relative importance of various weather variables for crop yield prediction.
Advances in developing an explainable yield prediction model using attention mechanism is an attractive development. 
The basic framework of LSTM for the phenotypic prediction can be applied to any crop with weather-dependent variability in order to better understand the genotype x environment effects found in the course of multi-environment testing. 
As such, this approach can be immediately useful for researchers in a variety of crops and environments and may prove to be exceptionally powerful when used in collaborative efforts between researchers operating in contrasting climatic zones, and in conjunction with sensor data for prescriptive breeding\cite{parmley2019scirep} including for root traits\cite{falk2020computer}.
The insights provided by our model can help in understanding the impact of weather variability on agricultural production in the presence of climate change, and devise breeding strategies for variety plasticity to circumvent these climatic challenges.

The ability to make accurate predictions of crop performance can lead to optimization across many different levels of organizations. At the federal level, improved crop insurance recommendations can be made based on weather forecasts before planting, and be continually updated throughout the season as more data is recorded and forecasts are updated. Railroads, grain cooperatives, and end-users can streamline the logistics of handling the desired quantities of grain if they are permitted a better understanding of how much grain (and of what quality) will be produced in a given region. Farmers can make better marketing decisions if they have an accurate and high confidence prediction of their production for the year, allowing them to sell their crops at the most opportune time. We envision that similar work on other crops and over a longer time span will generate invaluable insights for cultivar development and plant breeding and production related research in a challenging climate. 


\section*{Conclusion}
Unraveling causality would be a substantial step forward in understanding impact of climate change on variety's plasticity. Viewed through the lens of causality, DL based predictive models vs process based predictive models have distinct pros and cons. Process based models have clear causal relationships (by construction); however causality is limited to the confines of the model parameters, and it is non-trivial to assimilate additional data to extract broader causal trends. 
On the other hand, incorporating causality into DL based models is an open problem in the AI/ML community, with much activity. No principled approaches exist to accomplish this. However, DL based models (in contrast to process-based models) have the ability to seamlessly assimilate additional data. Our vision is therefore to evaluate if systematically augmenting DL based predictive models with increasing amounts of  physio-morphological informative features provides a way towards unraveling causal relationships.
We accomplish this by deploying our DL framework as a 'hypotheses generation tool'. We build DL models using a large volume of data and variety of information incorporating domain based knowledge. We then systematically probe the impact of various physio-morphological and environmental parameters on yield (via sensitivity analysis, and "what if" scenario evaluation), and  establish a framework to generate hypotheses in different crop species and physio-morphological characteristics under different climatic conditions. 
Until causality based DL becomes feasible, the hypotheses generation DL models will have the maximum impact in meeting the need of climate change scenarios and to incorporate plasticity response in future varieties.

\section*{Acknowledgements}

Funding for this project was provided by Iowa Soybean Association (AKS), Monsanto Chair in Soybean Breeding (AKS), RF Baker Center for Plant Breeding (AKS), Plant Sciences Institute (SS, BG and AKS), USDA (SS, BG, AKS), NSF NRT (graduate fellowship to JS) and ISU's Presidential Interdisciplinary Research Initiative (AKS, BG, SS). The authors thank Vikas Chawla for his assistance with querying weather data for this project.



\section*{Author contributions statement}

A.K.S., J.S., S.S. and B.G. conceived the research; All authors contributed in the design of the analysis and interpretation; J.S. compiled the UST performance and pedigree data; T.G. and J.S. performed statistical analysis, T.G. and L.W. built machine learning models and results were interpreted by T.G. and J.S. with inputs from S.S., A.K.S., and B.G.; J.S. and T.G. wrote the first draft with inputs from A.K.S. and S.S.; All authors contributed to the development of the manuscript.





\bibliography{bib.bib}

\clearpage

\section*{Supplementary Materials}

\section*{Clustering}
Clustering is an unsupervised machine learning technique used to group unlabeled examples. A metric (similarity measure) is used to estimate the similarity between examples by combining the examples' feature data. With the increase in the number of features, the similarity measure computation can become more complex. By assigning a number to each cluster, each complex example is represented by a cluster-ID. This makes clustering a simple yet powerful technique that finds applications in domains including image segmentation, anomaly detection, social network analysis, and medical imaging. The output of the clustering technique (Cluster ID) can be then used as input instead of a high-dimensional feature for machine learning algorithms. 

The choice of a clustering algorithm depends on whether it can scale efficiently to the available dataset. The clustering algorithms that compute the similarity between all pairs of examples, are not practical to be used for a large number of examples ($n$) as the runtime for this type of algorithm is proportional to the square of $n$. K-means clustering algorithm scales linearly with $n$ and thus can be used for large-scale data. While the centroid-based clustering algorithm organizes data into non-hierarchical clusters, density-based clustering works by connecting highly dense areas into clusters. Density-based clustering is not suitable for data with high-dimensions and distribution-based clustering is not applicable when the data-distribution type is not known. Hierarchical clustering, which computes a tree of clusters is mostly meant for hierarchical data. This leads to the choice of a centroid-based algorithm for clustering the genotypes represented by a correlation matrix. K-means clustering is simple, efficient and the most commonly used centroid-based clustering algorithm. We implemented the K-means clustering algorithm in this work for these reasons. 

\section*{Modeling Approach}
\subsection* {Long Short Term Memory Networks (LSTMs)}
Recurrent Neural Networks (RNNs) can explicitly capture temporal correlations in time series data, and efficient learning of the temporal dependencies leads to highly accurate prediction and forecasting, often outperforming static networks. Deep RNNs are trained using the error backpropagation algorithm; however, the propagation of error gradients through the latent layers and unrolled temporal layers suffer from the vanishing gradient problem. Therefore, gradient descent of an error criterion may be inadequate to train RNNs especially for tasks involving long-term dependencies \cite{bengio1994learning}. Moreover, standard RNNs fail to learn in the presence of time lags greater than 5-10 discrete time-steps between relevant input events and target signals ~\cite{gers1999learning}. To overcome these challenges, Long short-term memory (LSTM) was used, which is an RNN architecture designed to overcome the error backflow problems \cite{hochreiter1997long}. By using input, output and forget gates to prevent the memory contents being perturbed by irrelevant inputs and outputs, LSTM networks have the ability in learning long-range correlations in a sequence and can accurately model complex multivariate sequences~\cite{malhotra2015long}.  

\begin{figure*}[t]
\begin{center}
\includegraphics[width=10cm, keepaspectratio]{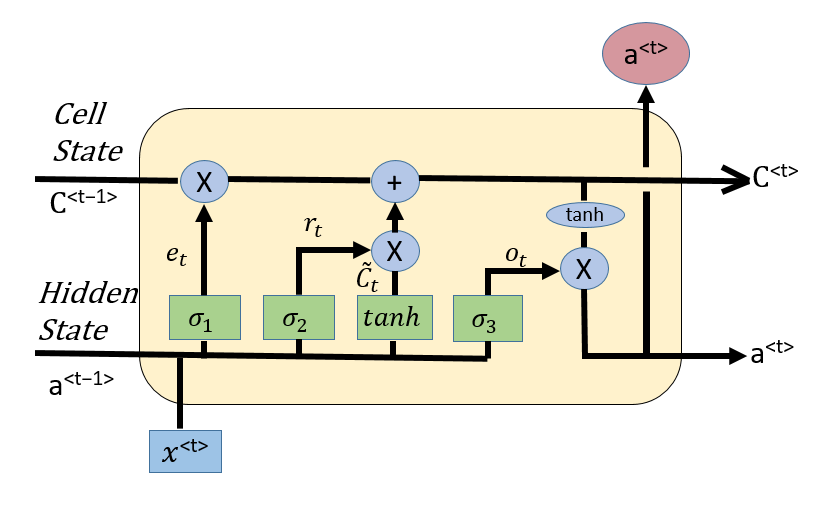}
\end{center}
\caption{An overview of the LSTM block. The input, output and forget gates regulate whether information can be augmented or removed from the cell state.}
\label{lstm_block}
\end{figure*}

The cell state in an LSTM block can allow the information to just flow along with it unchanged and information can be added to or removed from the cell state.
In an LSTM block, there are input, output and forget gates that prevent the perturbation of the memory contents with irrelevant information. 
These gates regulate the augmentation of any information to the cell state. An overview of the LSTM block is demonstrated in Fig.~\ref{lstm_block}. 

Forget gate layer is the first step of a LSTM block. A sigmoid layer ($\sigma_1$) decides the information to be removed the cell state of the previous time-step $C^{<t-1>}$. Forget gate naturally permits LSTM to learn local self-resets of memory contents that have become irrelevant ~\cite{gers1999learning}. This is performed as:

\vspace{-0.2cm}
\begin{equation}
e_t = \sigma_{1}(W_e \cdot [a^{<t-1>},x^{<t>}] + b_e )\\
\end{equation}

The next step is to augment the cell state with new information. A sigmoid layer($\sigma_2$, the input gate layer) followed by a tanh layer generates potential new information $\tilde C^{<t>}$ for augmentation.  

\vspace{-0.2cm}
\begin{equation}
r_t = \sigma_{2}(W_r \cdot [a^{<t-1>},x^{<t>}] + b_r )\\
\tilde C^{<t>} = tanh(W_C \cdot [a^{<t-1>},x^{<t>}] + b_C )\\
\end{equation}

Thereafter, the new cell state $C^{<t>}$ is obtained as follows:

\vspace{-0.2cm}
\begin{equation}
C^{<t>} = e_t \times C^{<t-1>} + r_t \times \tilde C^{<t>}\\
\end{equation}

Thereafter, the hidden state of the previous time-step $a^{<t-1>}$ is passed through the third sigmoid layer ($\sigma_3$) for information selection. After combining with the new cell state $C^{<t>}$ (filtered with tanh layer), the new hidden state $a^{<t>}$ is computed as:

\vspace{-0.2cm}
\begin{equation}
o_t = \sigma_{3}(W_o \cdot [a^{<t-1>},x^{<t>}] + b_o )\\
a^{<t>} = o_t \times tanh(C^{<t>})\\
\end{equation}

After updating, the values of $a^{<t>}$ and $C^{<t>}$ are passed to the LSTM block of the next time-step. The forget gate and output activation function is the most critical components of the LSTM block and removing any of them can significantly impair performance ~\cite{greff2017lstm}.

\subsection* {Encoding Using Stacked LSTM}
\begin{figure*}[t!]
\begin{center}
\setlength{\unitlength}{0.012500in}%
\includegraphics[width=15cm]{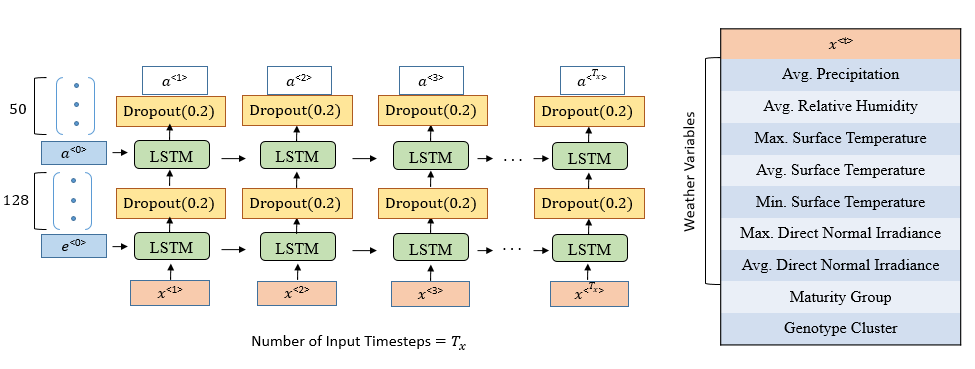}
\end{center}
\caption{LSTM is used for encoding the input sequence which is of length $T_x$ and the output from the first LSTM layer is a batch of sequences that are propagated through another layer of LSTM. We used dropout regularization after each LSTM layer to prevent overfitting.}
\label{lstm_encoding}
\end{figure*}

To encode the information of the input time-steps, we had two LSTM layers stacked on top of each other to get the $T_x$ annotations as shown in Fig.~\ref{lstm_encoding}. We finalized this model after an extensive hyper-parameter and architecture search.
The Stacked LSTM can capture the long-range dependencies and temporal correlations for nonlinear data, therefore, they were ideal for our research problem.
An LSTM layer consists of a sequence of directed nodes where each node corresponds to a single time-step.


The first layer of LSTM takes in input the information from all timesteps sequentially. For input at each time-step, we concatenated the information from different variables and this concatenated information acts as the input to the LSTM node. The LSTM node computes the hidden state as a function of the previous hidden state and the input vector for the current timestep. Each LSTM node updates the hidden state and cell state. The next LSTM node receives as input these updated states and the concatenated information of that time-step. After performing computations at each timestep of the time series, the first LSTM layer generates a sequence of hidden states for input to the next LSTM layer. The encodings returned by the first LSTM layer act as inputs for the next LSTM layer.

\subsection* {Temporal Attention}
\begin{figure*}[t!]
\begin{center}
\setlength{\unitlength}{0.012500in}%
\includegraphics[width=10cm]{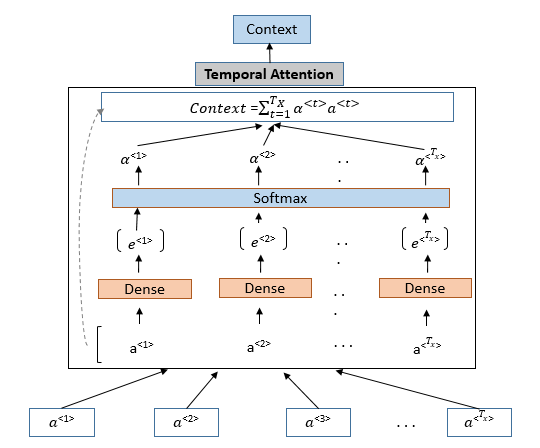}
\end{center}
\caption{The temporal attention method used to compute the context vector and learn the attention weights simultaneously.}
\label{temporal_attention}
\end{figure*}

The temporal attention method takes in input a sequence of vectors and the aim is to compute aggregated information from these vectors. The vectors are annotations corresponding to the input time-steps. We computed the context vector from the weighted sum of annotations (hidden states) as shown in Fig.~\ref{temporal_attention}. Annotation  $a^{<t>}$ focuses on the information surrounding the time-step $t$ in the sequence. The attention weight $\alpha^{<t>}$ signifies the contribution of the information at a time-step $t$ for prediction. 

The context vector is computed like this:

\vspace{-0.5cm}
\begin{equation}
context = \sum_{t = 1}^{T_x} \alpha^{<t>}a^{<t>}\\
\end{equation}

The attention weight for each annotation $a^{<t>}$ was computed using the softmax function. 

\vspace{-0.5cm}
\begin{equation}
\alpha^{<t>} = \frac{exp(e^{<t>})}{\sum_{t=1}^{T_x} exp(e^{<t>})}\\
\end{equation}

The alignment model (dense layer, $d$) scores how well the input around time-step $t$ is aligned with the prediction. It is parameterized as a feedforward neural network model as shown in Fig.~\ref{temporal_attention}. It is jointly trained with the entire network.  

\vspace{-0.5cm}
\begin{equation}
e^{<t>} = d(a^{<t>})\\
\end{equation}

\section*{Experiments}
\subsection*{Downsampling the dataset}
The original multivariate time-series data set comprises of 214 days (thus having 214 time-steps). Each day is represented by 7 weather variables (Table~\ref{Table: Weather Variables}). We initially perform experiments using our Stacked LSTM model keeping the number of time-steps ($T_x$) as 214. We vary the number of input time-steps to compute the effect $T_x$ has on the accuracy of the prediction.  We consider the first 210 time-steps while downsampling the daily data to weekly ($T_x=30$), biweekly ($T_x=15$) and monthly ($T_x=7$) values. We kept the sense of the variables the same while downsampling and thus instead of considering mean for all the variables, we compute an average of average (ADNI, ARH, AvgSur), maximum of maximum (MDNI, MaxSur), minimum of minimum (MinSur). For average precipitation, we consider both the total precipitation and the average precipitation for the considered time interval (7 days, 14 days, 30 days). The model performs better when the downsampling is performed using average precipitation and therefore we implement this in our experiments. \footnote{The abbreviations mentioned in Table~\ref{Table: Weather Variables} are used in Table~\ref{Table: greedy_search_weather_variables}, Table~\ref{Table: greedy_search_north} and Table~\ref{Table: greedy_search_south}.}

\begin{table}[!hbt]
\centering
\caption{Abbreviations used for different weather variables. At each time-step of the LSTM architecture, we used multivariate input.}
\label{Table: Weather Variables}
\begin{tabular}{|c|c|c|}
\hline
Weather Variable & Explanation & Unit \\ \hline
ADNI & Average Direct Normal Irradiance & $Wm^{-2}$ \\ \hline
AP   & Average Precipitation Previous Hour & $inches$ \\ \hline 
ARH  & Average Relative Humidity & $Percentage$ \\ \hline
MDNI  &  Maximum Direct Normal Irradiance & $Wm^{-2}$ \\ \hline 
MaxSur  &  Maximum Surface Temperature & \textdegree$C$ \\ \hline MinSur  &  Minimum Surface Temperature & \textdegree$C$ \\ \hline
AvgSur  &  Average Surface Temperature & \textdegree$C$ \\ \hline
\end{tabular}
\end{table}

\subsection*{Number of Input time-steps}
We kept the model architecture the same while performing experiments with different input sequence lengths. The RMSE values corresponding to different $T_x$ were almost same as observed in Table~\ref{Table: Varying Input Sequence Length} except for $T_x = 7$. For almost all of the experiments, we chose an intermediate value of $T_x = 30$ to facilitate faster training of LSTMs, to capture long-range temporal dependencies and also, not downsampling to a higher extent.

\begin{table}[!hbt]
\centering
\caption{Test Set RMSE values for different input sequence lengths.}
\label{Table: Varying Input Sequence Length}
\begin{tabular}{|c|c|}
\hline
Input Sequence Length ($T_x$)  & Test RMSE \\ \hline
7  & 7.216 \\ \hline
15 & 7.205 \\ \hline
30 & 7.207 \\ \hline
214 & 7.206 \\ \hline
\end{tabular}
\end{table}

\subsection*{Adding Maturity Group, Genotype Cluster Informations}
We performed three different experiments to optimize the augmentation of the information of Maturity Group (MG). To do this, we developed three different architectures. In the first approach, we concatenated MG to every time-step of the input. In the second approach, we concatenated MG to every time-step and also after the 2nd LSTM layer just before prediction. For the third architecture, we added MG only after the 2nd LSTM. The results are given in Table~\ref{Table: adding_mg}. The second approach which gave the least RMSE became our selected approach for the paper. Interestingly, we observe that even though MG is a static value for all time-steps, approach 1 gave better results than approach 3.  

\begin{table}[!hbt]
\centering
\caption{Different approaches to augment MG information.}
\label{Table: adding_mg}
\begin{tabular}{|c|c|}
\hline
Approach  & Test RMSE \\ \hline
MG added to every TS  & 7.246 \\ \hline
MG added to every TS \& after 2nd LSTM & 7.223 \\ \hline
MG added only after 2nd LSTM & 8.138 \\ \hline
\end{tabular}
\end{table}

With MG information already augmented, we performed a similar search to select a suitable architecture for adding the genotype cluster information. We observed that approach 2 also gives the least RMSE in this case (Table~\ref{Table: adding_cluster}). 

\begin{table}[!hbt]
\centering
\caption{Different approaches to augment genotype cluster information.}
\label{Table: adding_cluster}
\begin{tabular}{|c|c|}
\hline
Approach  & Test RMSE \\ \hline
Cluster added to every TS  & 7.230 \\ \hline
Cluster added to every TS \& after 2nd LSTM & 7.201 \\ \hline
Cluster added only after 2nd LSTM & 7.239 \\ \hline
\end{tabular}
\end{table}

\subsection*{Greedy Search - Weather Variables}
With MG and cluster information augmented, we perform a greedy search to get insights about which weather variable is most important for prediction. In the first step of the greedy search, we trained our Stacked LSTM model for each of the 7 variables listed in Table~\ref{Table: Weather Variables} separately. We chose the variable which gave the least RMSE value. With MinSur as the chosen first variable, in the second step, we trained 6 models with each of the other 6 variables added. Thereafter, we selected the variable which gave the lowest RMSE and thus, ADNI becomes the second most important variable. The third step involved training 5 models and we repeated this greedy search process till we got our 7th variable. The variables in decreasing order of importance (as obtained in the greedy search) are listed in Table~\ref{Table: greedy_search_weather_variables}. We observed that the RMSE values remain in the range (7.19 - 7.20) and adding extra weather variables after the 1st variable didn't show a considerable impact on the RMSE values.    

\begin{table}[!hbt]
\centering
\caption{Greedy Search for the weather variables (Entire dataset).}
\label{Table: greedy_search_weather_variables}
\begin{tabular}{|c|c|}
\hline
Variable Added  & Test RMSE \\ \hline
MinSur  & 7.204 \\ \hline
ADNI & 7.202 \\ \hline
AvgSur & 7.207 \\ \hline
MDNI & 7.190 \\ \hline
AP & 7.189 \\ \hline
ARH & 7.211 \\ \hline
MaxSur & 7.203 \\ \hline
\end{tabular}
\end{table}

We performed a similar greedy search to get insights about the significance of variables separately for the northern and the southern parts. To do this, we divided our dataset based on the locations into two parts - North (MG 0 to 4) and South (MG 4 to 8). Our greedy search results for north data are shown in Table~\ref{Table: greedy_search_north}, while for the south data are shown in Table~\ref{Table: greedy_search_south} in decreasing order of importance for the variables. 

\begin{table}[!hbt]
\centering
\caption{Greedy Search for the weather variables (northern locations).}
\label{Table: greedy_search_north}
\begin{tabular}{|c|c|}
\hline
Variable Added  & Test RMSE \\ \hline
ADNI  & 7.206 \\ \hline
MDNI & 7.202 \\ \hline
MinSur & 7.206 \\ \hline
ARH & 7.208 \\ \hline
MaxSur & 7.224 \\ \hline
AvgSur & 7.210 \\ \hline
AP & 7.228 \\ \hline
\end{tabular}
\end{table}

\begin{table}[!hbt]
\centering
\caption{Greedy Search for the weather variables (southern locations).}
\label{Table: greedy_search_south}
\begin{tabular}{|c|c|}
\hline
Variable Added  & Test RMSE \\ \hline
AvgSur  & 7.233 \\ \hline
MDNI & 7.236 \\ \hline
MaxSur & 7.222 \\ \hline
ARH & 7.234 \\ \hline
ADNI & 7.223 \\ \hline
MinSur & 7.249 \\ \hline
AP & 7.265 \\ \hline
\end{tabular}
\end{table}

\subsection*{Performance Comparison of two models with RF, LASSO}
We compared the performance of the two models (Stacked LSTM, Temporal Attention) with Random Forest (RF) and LASSO regression on same input features. RF and LASSO models were developed using scikit-learn Python library. We vary the input sequence length $T_x$ to demonstrate the comparative results for these models. Each model has two variants based on the input information. For one variant, the input included 7 weather variables and for the other variant, the input included 7 weather variables, MG and cluster. The two models demonstrated comparable accuracy as shown in Table~\ref{Table: results_compare}. The best RMSE value (7.130) obtained for $T_x = 30$ is about 14\% of the average yield for the test set (50.745) and 44.5\% of the standard deviation (16.019). Both the Stacked LSTM and Temporal Attention models outperformed RF and LASSO. 

\begin{table}
\centering
\caption{Comparison of the performance of the two models (2 variants of each model based on the input information) with Random Forest and LASSO by varying the input sequence length ($T_x$) using test RMSE values.}
\label{Table: results_compare}
\begin{threeparttable}
\begin{tabular}{|c|c|c|c|} 
\hline
$T_x$ & Model &  RMSE(Weather Variables) &  RMSE(Including all)\\ \hline
 7  & Temporal Attention & 8.272 & 7.137\\
  & Stacked LSTM & 8.246 & 7.132\\ 
  & Random Forest & 11.146 & 11.052\\
  & LASSO & 14.466 & 14.444\\
\hline
 \midrule
 15  & Temporal Attention & 8.254 & 7.129\\
  & Stacked LSTM & 8.251 & 7.131\\ 
  & Random Forest & 11.018 & 10.666\\
  & LASSO & 13.732 & 13.711\\
\hline
  \midrule
 30  & Temporal Attention & 8.236 & 7.148\\
  & Stacked LSTM & 8.289 & 7.130\\ 
    & Random Forest & 10.131 & 9.889\\
  & LASSO & 12.813 & 12.779\\
\hline

 \bottomrule
\end{tabular}
\end{threeparttable}
\end{table}

\subsection*{Results with different inputs to the Stacked LSTM model}
For the Stacked LSTM model, we vary the inputs to the model in order to notice the change in performance. As both of our proposed models showed comparable accuracy, we chose only one model to do these experiments. The results are shown in Table~\ref{Table: results_diff_inputs} in terms of RMSE and $R^2$ (coefficient of determination) regression score function for the test set. All the information from MG, cluster and weather variables were required to get the best performance.  

\begin{table}[!hbt]
\centering
\caption{Comparative performance of the Stacked LSTM for different inputs.}
\label{Table: results_diff_inputs}
\begin{tabular}{|c|c|c|}
\hline
Model Input  & Test RMSE & Test Set $R^2$ \\ \hline
Only MG  & 15.201 & 0.099\\ \hline
Only Cluster & 15.612 & 0.050\\ \hline
Only Weather Variables & 8.274 & 0.733\\ \hline
MG, Cluster & 15.149 & 0.106\\ \hline
MG, Weather Variables & 7.211 & 0.797\\ \hline
Cluster, Weather Variables & 8.159 & 0.741\\ \hline
MG, Cluster \& Weather Variables & 7.130 & 0.802\\ \hline
\end{tabular}
\end{table}

\subsection*{Comparison with USDA Model}
We compare the performance of our deep learning model (Stacked LSTM) with the USDA's weather-based soybean yield prediction model \cite{jewison2013USDA}. The USDA model which uses a linear regression approach doesn't predict performance for individual locations. It predicts yield state-wise. Due to this limitation of the USDA model, we compare the models using year wise average across states for the test set (Table~\ref{Table: results_usda_comparison}). We computed the absolute error (between predicted and actual yield) for both the models. The deep learning model showed much-improved performance compared to the domain knowledge-based USDA model. 


\begin{table}[!hbt]
\centering
\caption{Comparative performance of Deep Learning (Stacked LSTM) model with the USDA model.}
\label{Table: results_usda_comparison}
\begin{tabular}{|c|c|c|}
\hline
Year & Absolute Error (Deep Learning) & Absolute Error (USDA) \\ \hline
2003  & 0.63 & 5.49\\ \hline
2004  & 0.12 & 0.33\\ \hline
2005  & 0.19 & 2.75\\ \hline
2006  & 0.44 & 2.18\\ \hline
2007  & 0.21 & 0.37\\ \hline
2008  & 0.30 & 2.98\\ \hline
2009 & 0.04 & 0.84\\ \hline
2010 & 0.40 & 1.37\\ \hline
2011 & 0.34 & 0.06\\ \hline
2012 & 0.29 & 1.89\\ \hline
2013 & 0.04 & 0.27\\ \hline
2014 & 0.03 & 1.32\\ \hline
2015 & 0.35 & 1.70\\ \hline
\end{tabular}
\end{table}


\subsection*{Data Availability - Insights}
\begin{figure*}[t!]
\begin{center}
\setlength{\unitlength}{0.012500in}%
\includegraphics[width=16cm]{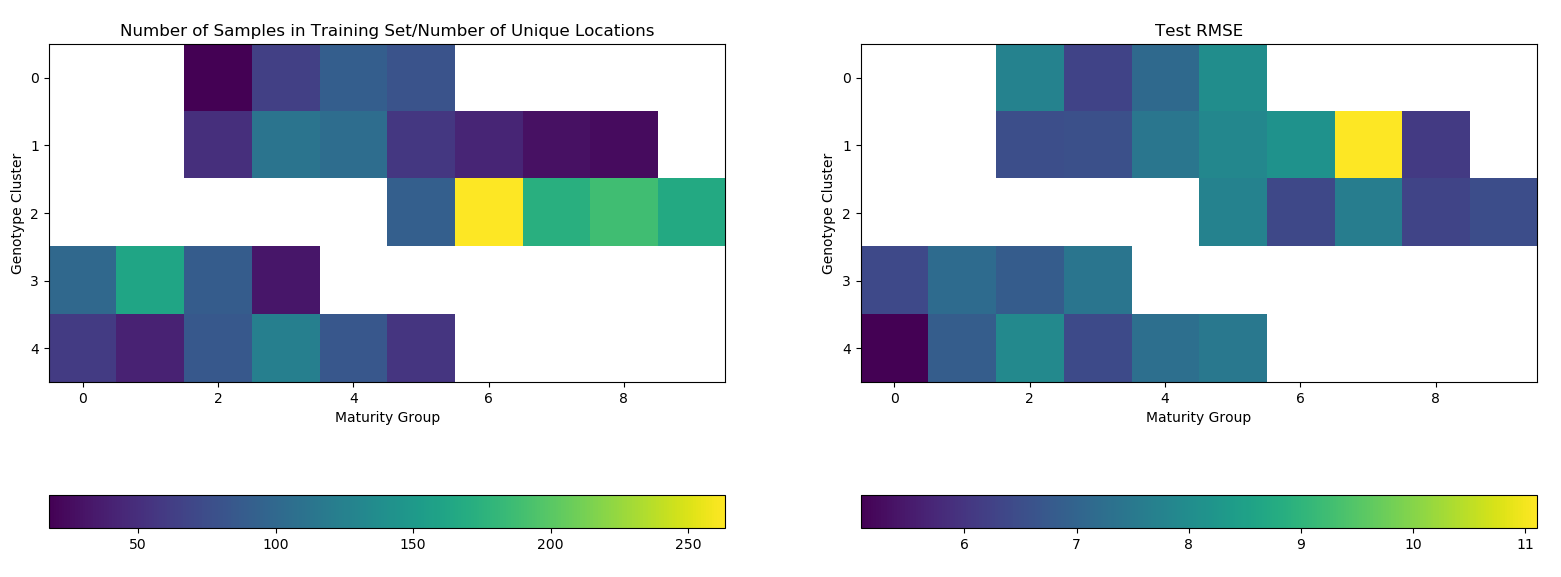}
\end{center}
\caption{Attempt to gain insights behind performance on the test set based on data availability in the training set.}
\label{data_availability}
\end{figure*}

We try to gain insights based on data availability for performance on the test set as shown in Fig ~\ref{data_availability}. We plotted the test RMSE values using a heat map for all the (MG, Genotype Cluster) combinations. We also plotted the ratio (number of samples in training set/number of unique locations) to get an estimation of the data availability and data distribution for all the (MG, Genotype Cluster) combinations. From the figure, we observed that for the highest RMSE (MG = 7, Cluster = 1) the corresponding data availability ratio is low. This holds true for most of the highest RMSE combinations, although not for all. Therefore, while not conclusive, it seems bad performance can be attributed to less data availability/location in the training set leading to high RMSE in the predicted yield. The framework developed in this paper allows similar investigations to motivate other insights, and future hypotheses driven research.  

\pagebreak

\end{document}